\pdfoutput=1

\documentclass[11pt]{article}

\usepackage{acl}
\usepackage{flushend}

\usepackage{times}
\usepackage{latexsym}
\usepackage{xcolor}
\definecolor{Mycolor2}{HTML}{FCC267}
\definecolor{Mycolor1}{HTML}{AECDC2}
\usepackage{multirow}
\usepackage{amsfonts}
\usepackage{colortbl}
\usepackage{cellspace}
\usepackage[T1]{fontenc}

\usepackage[utf8]{inputenc}

\usepackage{microtype}

\usepackage{comment}
\usepackage{booktabs}
\usepackage{xcolor}
\usepackage[inline]{enumitem}
\usepackage{amsmath}
\usepackage{graphicx} 
\usepackage{caption}

\usepackage{array}
\usepackage{makecell}

\usepackage{amsmath}

\usepackage{flushend}

\title{Measuring Geographic Performance Disparities of\\Offensive Language Classifiers}

\author{Brandon Lwowski$^1$, Paul Rad,$^{1,2}$ and Anthony Rios$^{1}$ \\
  $^1$Department of Information Systems and Cyber Security\\
   $^2$Department of Computer Science \\
  University of Texas at San Antonio \\
  \texttt{\{brandon.lwowski, peyman.najafirad, anthony.rios\}@utsa}}

\begin{document}

\maketitle

\begin{abstract}
Text classifiers are applied at scale in the form of one-size-fits-all solutions. Nevertheless, many studies show that classifiers are biased regarding different languages and dialects. When measuring and discovering these biases, some gaps present themselves and should be addressed. First, ``Does language, dialect, and topical content vary across geographical regions?'' and secondly ``If there are differences across the regions, do they impact model performance?''. We introduce a novel dataset called GeoOLID with more than 14 thousand examples across 15 geographically and demographically diverse cities to address these questions. We perform a comprehensive analysis of geographical-related content and their impact on performance disparities of offensive language detection models. Overall, we find that current models do not generalize across locations. Likewise, we show that while offensive language models produce false positives on African American English, model performance is not correlated with each city's minority population proportions.
\textbf{Warning: This paper contains offensive language.}
\end{abstract}

\section{Introduction}


Many tasks revolving around text classification of social network data have been introduced including, but not limited to tracking viruses~\cite{lamb2013separating,corley2009monitoring, corley2010text, santillana2015combining, ahmed2018social, lwowski2020covid}, providing help for (natural) disasters~\cite{neubig2011safety,castillo2016big,reuter2018fifteen}, detecting misinformation~\cite{oshikawa2020survey}, and identifying cyber-bullying~\cite{xu2012learning}. Overall, text classifiers have been shown to be ``accurate'' across a wide range of applications. As deep learning models and packages have made substantial progress for the field of natural language processing (NLP), NLP models have become more accessible to the general public. Hence, models are being deployed in a production environment and run at scale at a growing pace. However, recent work has shown that these models are biased and unfair, especially towards minority groups~\cite{blodgett2016demographic,davidson2019racial}. In this paper, we expand on prior work by analyzing how model performance can fluctuate due do geographically-caused differences in language and topical content that exists in the context of offensive language detection.


Researchers have shown that topical and stylistic attributes of text are used by speakers on social media to implicitly mark their region-of-origin~\cite{shoemark2017topic, hovy2018capturing, cheke2020understanding, gaman2020report}. For instance, \citet{hovy2018capturing} show that doc2vec embedding frameworks can be leveraged to detect geolocation-related language differences. \citet{hovy2020visualizing} then introduces visualization techniques for measuring regional language change. \citet{kellert2021geolocation} shows that differences exist at the city level as well.
Hence, prior work has generally focused on incorporating or identifying regional aspects of language data to improve performance in machine translation~\cite{ostling2017continuous} or geolocation prediction and clustering~\cite{hovy2018capturing}.

For particular downstream tasks, recent work in understanding performance disparities has found differences across various languages~\cite{gerz2018relation} (e.g., Finish vs. Korean) and dialects~\cite{davidson2017automated,sap2019risk}---such as African American English (AAE).  Likewise, \citet{davidson2019racial} and \citet{sap2019risk} show that abusive and hate speech-related language classifiers are biased against AAE-like text and machine learning models can learn these biases when certain populations are not being represented, making the data unbalanced. These results have been shown to extend into other text classifications tasks, for example, \citet{lwowski2021risk} show that influenza detection models are also biased against AAE-like text. Similarly, \citet{hovy2015tagging} find that part-of-speech tagging performance correlates with age. 



While there has been a substantial amount of research understanding, identifying, and measuring performance disparities across languages and dialects, to the best of our knowledge, there has been no prior work on measuring the performance of NLP classifiers across different geographic regions. Specifically, prior work has not measured how geographical variations in language and topical content---or stance towards certain topics---impacts the performance of offensive language classifiers. Complex interactions between topical content and style can impact model performance.

In the context of AAE-related studies~\cite{sap2019risk}, AAE is not spoken the same across different regions of the United States. There have been multiple studies in diversity, equity, and inclusion arguing against treating African Americans as a monolithic group of people~\cite{tadjiogueu2014fifty, erving2021disrupting}. Moreover, certain features of AAE only appear within specific regions of the US~\cite{jones2015toward}. Likewise, geographic factors have been known to impact social behaviors such as voting turnout~\cite{zingher2019power} and general health disparities~\cite{thomas2014overcoming}. Hence, these geographic factors can impact both \textit{how} people write and \textit{what} people write about on social media. Hence, to start addressing these issues, this paper proposes an initial study looking at how individual offensive language model's performance can vary geographically for the task of detecting offensive language due to the stylistic  and topical differences in language.



Overall, to better understand the implications of geographical performance disparities offensive language models, we make three contributions: 
\begin{enumerate}[label=\bfseries (\arabic*.), topsep=0pt]
\itemsep-.6em 
\item To the best of our knowledge, we perform the first analysis of geographical performance variation of offensive language classification models, producing novel insights and a discussion of important avenues of future research.
\item We introduce a novel labeled offensive language dataset called GeoOLID with more than 14 thousand tweets across 15 geographically and demographically diverse cities in the United States.
\item We produce a comprehensive manual error analysis, grounding some performance disparities to stance and topics.
\end{enumerate}

\section{Language Variation}

To the best of our knowledge, the impact geographical variation in language style and topical content has not yet been studied in the context of offensive language detection to the best of our knowledge. 
Language variation is an important area of research for the NLP community. While there has been disagreement about whether morphology matters, \citet{park2021morphology} has shown that incorporating information that can model morphological differences is important in improving model performance. Prior work  has generally focused on either developing methods to identify language features within text or use various language features to improve model performance. Early work by \citet{bamman2014distributed} showed that embeddings can capture geographically situated language, while \citet{doyle2014mapping} explored ways to quantify regional differences against a background distribution. Recently, VarDial has hosted an annual competition to identify various dialects of different languages (e.g., German and Romanian) as well as geolocations~\cite{gaman2020report}.

\citet{cheke2020understanding} use topic distributions to show that different topics can provide signal to determine where the text originated from. For the same shared task, \citet{scherrer2021social} show that combining modern NLP architectures like BERT with a double regression model can also provide success in determining the latitude and longitude points of the location for the given text. The results of this shared task highlights the fact that topical and lexical differences exist based on the location a tweet was written. Other work around regional variation of language \cite{hovy2018capturing, hovy2020visualizing, kellert2021geolocation} further prove that these differences in dialect and lexical patterns are significant across geographies.

\section{Performance Disparities}


Performance disparities across languages and dialects recently have received attention in NLP. For example, recent research shows that performance drops in text classification models across different sub-populations such as gender, race, and minority dialects \cite{dixon2018measuring, park2018reducing, badjatiya2019stereotypical, rios2020fuzze,  lwowski2021risk, mozafari2020hate}. \citet{sap2019risk} measure the bias of offensive language detection models on AAE. Likewise, \citet{park2018reducing} measure gender bias of abusive language detection models and evaluate various methods such as word embedding debiasing and data augmentation to improve biased methods.  \citet{davidson2019racial} shows that there is racial and ethnic bias when identifying hate speech online and show that tweets in the black-aligned corpus are more likely to get assigned as hate speech. Overall, performance disparities have been observed across a wide array of NLP tasks such as detecting virus-related text~\cite{lwowski2021risk}, coreference resolution~\cite{zhao2018gender}, named entity recognition~\cite{mehrabi2020man}, and machine translation~\cite{font2019equalizing}.

\begin{table}[t]
\centering
\resizebox{\columnwidth}{!}{%
\begin{tabular}{lllll} \toprule
\textbf{} & \textbf{Non Offensive} & \textbf{Offensive} & \textbf{Total} & \textbf{MDE}  \\ \midrule
\textbf{OLID} & 9,460 & 4,640 & 14,100 & .014 \\ \midrule \midrule
& \multicolumn{4}{c}{\textbf{Filtered/Unfiltered GeoOLID Dataset}} \\ \midrule
& \textbf{Non Offensive} & \textbf{Offensive} & \textbf{Total}  & \textbf{MDE} \\ \midrule
\textbf{Unfiltered GeoOLID} & --- & ---  & 5,013,474 & --- \\ 
\textbf{GeoOLID} & 9,259 & 4,831 & 14,090  \\  \midrule
\textbf{City Name} & \textbf{Non Offensive} & \textbf{Offensive} & \textbf{Total} & \textbf{MDE} \\ \midrule
Baltimore, MD & 630 & 277 & 907 & .054\\
Chicago, IL & 676 & 326 & 1002 & .052 \\
Columbus, OH & 616 & 301 & 917 & .054 \\
Detroit, MI & 549 & 367 & 916 &  .053 \\
El Paso, TX & 502 & 404 & 906 & .055 \\
Houston, TX & 635 & 297 & 932 & .054 \\
Indianapolis, IN & 600 & 307 & 907 & .055 \\
Los Angeles, CA & 660 & 298 & 958 & .053 \\
Memphis, TN & 564 & 368 & 932 & .054  \\
Miami, FL & 726 & 216 & 942 & .054 \\
New Orleans, LA & 607 & 325 & 932 & .054 \\
New York, NY & 717 & 265 & 982 & .053 \\
Philadelphia, PA & 629 & 337 & 966 & .054 \\
Phoenix, AZ & 577 & 355 & 932 & .054 \\
San Antonio, TX & 572 & 387 & 959 & .053 \\ \bottomrule
\end{tabular}%
}
\caption{Dataset Statistics.\vspace{-2em}}
\label{tab:datasets}
\end{table}


Overall, the major gap in prior work investigating language variation is that there has not been any studies evaluating the impact regional language has on the performance of downstream tasks, particularly offensive language detection. Hence, we measure performance disparities across geographical regions for the task of detecting offensive language. Furthermore, many groups that are studied are ``monolithic'', such as male vs. female (using an unrealistic assumption of binary gender~\cite{rios2020quantifying}), or AAE which is not universally spoken in the same way within different cities in the US. For example, \citet{jones2015toward} show that many well-known AAE patterns (e.g., sholl, an nonstandard spelling of ``sure'') do not appear uniformly across the US. Likewise, the discussion topics can also change regionally. Hence, if an offensive language detection model performs poorly on one set of AAE patterns or topics that only appear in a particular region, it can impact that location much more than others.
Hence, we believe that fine-grain regional analysis is a better future avenue to understand the real-world impact of NLP models.

\section{Data Collection and Annotation}

\begin{figure}[t]
    \centering
    \includegraphics[width=.6\linewidth]{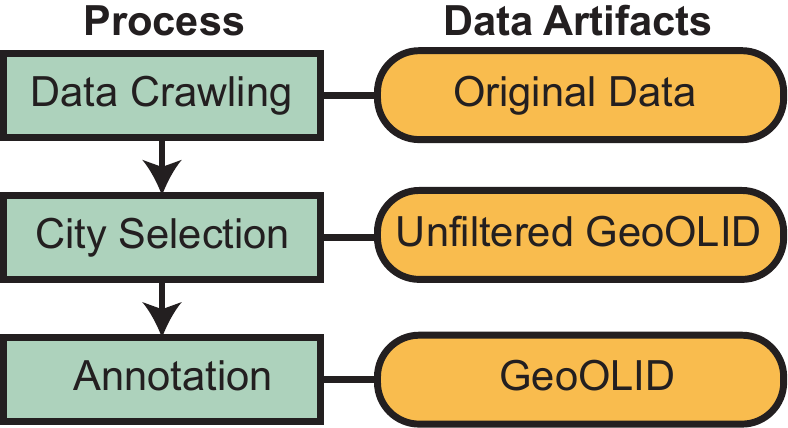}
    \caption{Data collection/annotation. Data collection steps are \colorbox{Mycolor1}{green} and produced datasets are in \colorbox{Mycolor2}{orange}. \vspace{-2em}}
    \label{fig:data-col}
\end{figure}


In this section, we describe the two major datasets used in our experiments: the Offensive Language Identification Dataset~(OLID)~\cite{zampieri2019predicting} and our newly constructed Geographical Diverse Offensive Language Identification Dataset (GeoOLID). A complete summary of the datasets can be found in Table~\ref{tab:datasets}. Furthermore, we provide a summary of the data collection and annotation pipeline in Figure~\ref{fig:data-col}.  Intuitively, we have three main steps: Data Crawling, City Selection, and Data Annotation. We save the data after each step to be used throughout parts of our analysis. We describe the OLID dataset and each step below.

\vspace{2mm} \noindent \textbf{OLID.}
The OLID dataset introduced by \citet{zampieri2019predicting} contains 14,100 tweets labeled to identify different levels of offensiveness including, but not limited to, Not Offensive, Offensive, Targeted Offense, and Not Targeted Offense. Furthermore, Targeted Offenses are sub-categorized as targeting an individual, group, or other. For this study, we use the first level: Not Offensive (9,460 Total) and Offensive (4,640 Total).

\vspace{2mm} \noindent \textbf{Step 1: Data Crawling (Original Data).}
In addition to the OLID dataset, we introduce a new offensive language dataset using tweets collected since the start of the Covid-19 pandemic. The data set was crawled by \citet{qazi_geocov19_sigspatial} and \citet{lamsal2021design}, collecting more than 524 million multilingual tweets across 218 countries and 47,000 cities between the dates of February 1, 2020 and May 1, 2020. The data collection started on February 1, 2020 using trending hashtags such as \#covid19, \#coronavirus,
\#covid\_19. See \citet{qazi_geocov19_sigspatial} for complete details. Given the large amount of politically divisive discourse, racist remarks, and social impact of Covid-19, the collection provides a unique testbed to understand geographic model variation. Particularly, where researchers are exploring analyzing geospatial patterns of Covid-related content on Twitter~\cite{stephens2020geospatial}. If the models perform differently across locations, the it is difficult to interpret the results. We refer to this complete Covid-19 data as ``Original Data''.

\vspace{2mm} \noindent \textbf{Step 2: City Selection (Unfiltered  GeoOLID).} To measure the performance difference across varying locations, we decided on 15 cities based on multiple facets, data availability, geographic diversity, and demographic diversity. In deciding which cities to use for our study we first selected cities from different parts of the United States (North, South, East, West). Next we wanted cities that varied in size and were also demographically different. In table \ref{tab:PCC}, the total populations (reported in the thousands) of the selected cities varies, ranging from around 400,000 to almost 9 million. 

We also wanted cities that varied demographically, particularly with regard to African American and Hispanic/Latino population proportions.\footnote{We choose these groups because they align with classes in the \citet{blodgett2016demographic} dialect classifier.} In Table~\ref{tab:OLID-results}, cities like Baltimore, Memphis, New Orleans, and Detroit were chosen due to the high proportion of African Americans populations while, Indianapolis and Columbus had high proportions of White Non-Hispanic residents. El Paso, San Antonio and Phoenix have a close proximity to the Mexico boarder and higher percentage of Latino and Hispanic residents, which is very different from Columbus, having a smaller number of African American and Hispanic residents. In addition, we selected cities where we knew residents could use very distinct accents and phonics like New York and New Orleans. Overall, by selecting the 15 cities in Table~\ref{tab:datasets}, we created a diverse dataset with multiple ethnicities, language styles, and topical differences. We refer to this unlabeled dataset as ``Unfiltered GeoOLID''. The basic stats of this dataset are available in Table~\ref{tab:datasets} in the row titled ``Unfiltered GeoOLID''.

\vspace{2mm} \noindent \textbf{Step 3: GeoOLID.} 
Similar to prior work, we need to sample a large number of offensive and non-offensive tweets from Unfiltered GeoOLID~\cite{zampieri2019predicting}. Hence, we filter Unfiltered GeoOLID using the following lexicons and keyword filters: the badword lexicon~\cite{von2009offensive}, hatebase lexicon~\cite{davidson2017automated}, offensive-related phrases used for the original OLID dataset~\cite{zampieri2019predicting} (``you are'', ``she is'', ``he is'', ``conservatives'', ``liberals'', ``MAGA'', and ``antifa''), and additional Covid-specific phrases we found to be correlated with potential discrimination in the dataset (``chinese'', ``china'', ``asia'', ``asian'', ``wuhan''). Along with the aforementioned filters, we randomly sampled a subset of tweets for annotation. The final counts of each city can be found in Table \ref{tab:datasets}. This dataset is referred to as ``GeoOLID''.

\vspace{1mm} \vspace{2mm} \noindent \textbf{Annotation.} Overall, we performed multiple rounds of annotation until a quality dataset was created. First, in order to provide accurate labels for this study, samples of tweets were assigned to three graduate students to be labeled as ``offensive'' or ``not-offensive'' using the base guidelines provided by \citet{zampieri2019predicting} for the the OLID dataset. A total of 20 students were recruited and given a stipend of \$100 for their time and effort. Several meetings were set up before labeling started to answer questions and address implications. We use the Offensive definition provided by \citet{zampieri2019predicting} is defined as \textit{tweets containing any form of non-acceptable language (profanity) or a targeted offense, which can be veiled or direct. This includes insults, threats, and posts containing profane language or swear words.}

Following general annotation recommendations for NLP~\cite{pustejovsky2012natural}, the annotation process was completed in three stages to increase the reliability of the labels across geographic regions. First, before assigning tweets, we assured every tweet was assigned to three graduate students for annotation, providing us with three independent labels for each tweet. We then calculated the agreement between annotators, resulting in a Fleiss Kappa of .47, indicating moderate agreement.

Second, we (the authors) of the paper manually---and independently---adjudicated (i.e., re-annotated) the labels of each student, correcting miss-annotated tweets that were not agreed on by all three annotators. Common issues found during the process were labels of ``Not Offensive'' for tweets with ad-hoc mentions of the ``Wuhan Virus'' and offensive content found in the hashtag. Specifically, based on the work by \citet{dubey2020resurgence}, we decided that mentions of ``Wuhan Virus'' and other related terms like ``China Flu'' and ``Kung Flu'' were deemed offensive as it fit into the category of an targeted offense, which can be veiled or direct.  The second round of agreement scores increased to .83 representing ``almost perfect agreement,''~\cite{landis1977measurement}.

To further ensure annotation quality, the authors went through the tweets once again discussing and correcting any final disagreements among the second round adjudications, forming the final dataset described in Table~\ref{tab:datasets}. After collecting and adjudicating the responses, the total number of Offensive tweets were 4,831 compared to 9,259 Not Offensive. We also report Minimum Detectable Effect (MDE)~\cite{card2020little} for Accuracy in Table~\ref{tab:datasets}. Specifically, use the Binomial Power Test, which assumes that samples are unpaired, i.e., the new model and baseline evaluation samples are drawn from the same data distribution but are not necessarily the same samples. The MDE numbers assume an accuracy of .75, which results in a significant difference between two models being around .05. We plot more potential MDE scores for different baseline numbers in the Appendix, Figure \ref{fig:acc-power}.



\section{Experiments}

In order to address and test whether performance disparities exist across geographic regions for offensive language classifiers, we ran multiple experiments. We analyzed performance across the 15 cities in the GeoOLID dataset. In the following subsections, we provide the details of our experiments and provide evidence supporting that our GeoOLID dataset is representative of the Unfiltered GeoOLID dataset and that offensive language classifier performance can vary by geolocation. In the final subsection, we explore the performance and language similarities across different geolocations that have similar demographics.



\subsection{Data Representation Evaluation}

\begin{table}[t]
\centering
\resizebox{.6\linewidth}{!}{%
\begin{tabular}{lrr}
\toprule
                 & \textbf{F1}   & \textbf{Acc.} \\ \midrule
\textbf{Stratified}               & .059 & .056     \\
\textbf{Uniform}             & .062 & .062     \\
\textbf{Prior}                 & .008 & .068     \\ \midrule 
\textbf{BoW}                & .430 & .380     \\
\textbf{POS}              & .410 & .356     \\
\textbf{Dialect}               & .374 & .366     \\ \midrule 
\textbf{POS +   Dialect}     & .419 & .357     \\
\textbf{BoW +   Dialect}      & \textbf{.436} & \textbf{.381}     \\
\textbf{BoW + POS +   Dialect}    &   .431 & .370         \\ \bottomrule
\end{tabular}%
}
\caption{Location prediction. The Accuracy, Macro Precision, Macro Recall, and Macro F1 reported are the results when trained on a sample of the Unfiltered GeoOLID and predicted on the labeled GeoOLID dataset.\vspace{-1em}}
\label{tab:city_pred}
\end{table}




In this section, we aim to measure how well the GeoOLID dataset matches the Unfiltered GeoOLID data from each city. Specifically, we want to ensure that patterns found in the unfiltered data are still present within our annotated GeoOLID sample. If patterns in the GeoOLID dataset are not in Unfiltered GeoOLID, it is hard to argue that the errors are location-specific. They could simply be caused by our data filtering strategy. 

\vspace{2mm} \noindent \textbf{Methods.} To measure how representative our sample is, we train a location prediction model. Given a tweet, the goal of the model is to predict the city in which the text was posted. To train the model we use two sets of features: Content Features and Stylistics Features. The content features are made up of the top 5000 unigrams in the Unfiltered GeoOLID dataset. It is also important to note that all of the GeoOLID tweets are removed from the Unfiltered GeoOLID dataset before processing.

We also explore two sets of style Features: Part-of-Speech and Dialect Features. Specifically, we use unigram, bigram, trigram POS features. Moreover, the dialect features are the probabilities returned from the dialect inference tool from \citet{blodgett2016demographic}. Given a tweet, the tool outputs the proportion of African-American, Hispanic, Asian, and White topics. 

Finally, we train a Random Forest classifier on the Unfiltered GeoOLID dataset and the results are reported using the labeled GeoOLID dataset as the test set. Hyperparameters are optimized using 10-fold cross-validation on the training data.  Because of the large size of the Unfiltered GeoOLID dataset, we sample a random subset of ~35k examples from the Unfiltered GeoOLID dataset to reduce the training cost. The goal is not to achieve the most accurate predictions, but to simply see if we can predict location much better than random. If a small completely random sample shows this, that is better then requiring all of the data. We also compare the results to three random sampling methods to measure the difference between random guessing and the trained model: Stratified, Uniform, and Prior. Stratified makes random predictions based on the distribution of the cities in the training data, Uniform predicts cities with equal proportions, and Prior always predicts the most frequent city.

\begin{table}[t]
\centering
\resizebox{\linewidth}{!}{%
\begin{tabular}{lrrrrr}
\toprule
                 & \textbf{AAS} & \textbf{HLS} & \textbf{Tot.} & \textbf{AA} & \textbf{H/L} \\ \midrule
\textbf{Baltimore}    & .168     &    .193     & 585    & 338 (57.7\%)      &  45 (7.8\%)        \\
\textbf{Chicago}      &         .147   &  .204 & 2,450  & 801 (32.7\%)       & 819 (33.4\%)         \\
\textbf{Columbus}     & .146    &     .201     & 905    & 259 (28.6\%)      & 70 (7.7\%)         \\
\textbf{Detroit}      & .196   &     .214     & 639    & 496 (77.7\%)      & 51 (8.0\%)         \\
\textbf{El Paso}         & .158  &    .227       & 678    & 25 (3.7\%)      & 551 (81.2\%)         \\
\textbf{Houston}     & .161  &      .205      & 2,304  & 520 (22.6\%)      & 1,013 (44.0\%)         \\
\textbf{Indianapolis} & .151      &   .194     & 887    & 248 (28.0\%)     & 116 (13.1\%)         \\
\textbf{Los Angeles}    & .144    &      .204    & 3,898  & 336 (8.6\%)       & 1,829 (47.0\%)          \\
\textbf{Memphis}     & .209       &   .220    & 633    & 389 (61.6\%)      & 62 (9.8\%)         \\
\textbf{Miami}       & .140  &     .175       & 442    & 57 (12.9\%)      & 310 (7.0\%)         \\
\textbf{New Orleans}    & .182       &    .197   & 383    & 208 (54.2\%)      & 31 (8.0\%)         \\
\textbf{New York City}       & .126    &     .182     & 8,804  & 1,943 (22.1\%)     &  2,490 (28.3\%)         \\
\textbf{Philadelphia} & .157      &    .204    & 887 & 248 (27.9\%)      & 116 (13.1\%)         \\
\textbf{Phoenix}      & .144   &    .208       & 1,608  & 125 (7.8\%)      & 661 (41.1\%)         \\
\textbf{San Antonio}     & .175    &     .222     & 1,434  & 102 (7.2\%)       & 916  (63.9\%)        \\ \midrule \midrule
\textbf{AA PCC} & \multicolumn{5}{c}{.565 (\textit{p} value: .028)} \\ 
\textbf{H/L PCC} & \multicolumn{5}{c}{.167 (\textit{p} value: .55)} \\ \bottomrule
\end{tabular}%
}
\caption{Pearson Correlation Coefficient (PCC) between the AAS and HLS and city populations. Populations reported in thousands and percentages are in parenthesis.\vspace{-0em}}
\label{tab:PCC}
\end{table}

\begin{table*}[t]
\centering
\resizebox{.85\textwidth}{!}{%
\begin{tabular}{lrrrrrrrrrrrrrrrr}
\toprule
       & \textbf{Bal} & \textbf{Chi} & \textbf{Col} & \textbf{Det} & \textbf{ElP} & \textbf{Hou} & \textbf{Ind} & \textbf{LA} & \textbf{Mem} & \textbf{Mia} &  \textbf{NO} & \textbf{NY} & \textbf{Phi} & \textbf{Pho} & \textbf{SA} & \textbf{AVG} \\ \midrule

\textbf{Stratified} & .279 & .328 & .306 & .367 & .431 & .348 & .364 & .303 & .363 & .253 & .322 & .296 & .376 & .394 & .410  & .343 \\
\textbf{Uniform} & .362 & .443 & .398 & .458 & .453 & .381 & .390 & .368 & .420 & .302 & .412 & .342 & .412 & .444 & .456 & .403 \\ \midrule \midrule
\textbf{Linear SVM} &  .661 &  .615 &  .650 &  .714 &  .658 &  .568 &  .611 &  .643 &  .702 &  .660 &  .708 &  .625 &  .680 &  .613 &  .680 &   .653 \\ \midrule
\textbf{BiLSTM}      &  .678 &  .623 &  .651 &  .725 &  .660 &  .591 &  .643 &  .642 &  .720 &  .666 &  .694 &  .624 &  .705 &  .637 &  .669 &  .662 \\
\textbf{CNN}        &  \textbf{.720} &  \textbf{.662} &  \textbf{.684} &  \textbf{.745} &  \textbf{.688} &  \textbf{.611} &  \textbf{.674} &  \textbf{.670} &  \textbf{.745} &  \textbf{.701} &  \textbf{.736} &  \textbf{.663} &  \textbf{.743} &  \textbf{.657} &  \textbf{.703} &  \textbf{.694} \\ 
\textbf{LSTM}        &  .653 &  .614 &  .633 &  .709 &  .638 &  .570 &  .624 &  .620 &  .701 &  .661 &  .680 &  .600 &  .686 &  .615 &  .650 &  .643 \\
\textbf{BERT}       &  . 601 &  .629 &  .641 &  .684 &  .661 &  .602 &  .621 & .642 &  .651 &  .614 &  .635 &  .593 &  .668 &  .607 &  .665 &   . 634 \\ \midrule
\textbf{AVG} & .663 & .629 & .652 & \cellcolor{blue!25}{.715} & .661 & \cellcolor{red!25}{.588} & .635 & .643 & .704 & .660 & .691 & .621 & .696 & .626 & .673 \\ \bottomrule
\end{tabular}}
\caption{F1 score for each model-city combination. The largest and smallest average values are shaded in \colorbox{blue!25}{blue} and \colorbox{red!25}{red}, respectively. The model with the \textit{largest} F1 (larger is better) is \textbf{bolded} for each city. Each city's shortname is as follows: Chicago~(Chi), Detroit~(Det), Baltimore~(Bal), El Paso~(ElP), Los Angeles~(LA),  Houston~(Hou), Columbus(~Col), Indianapolis~(Ind), Miami~(Mia), Memphis~(Mem),  New York City~(NYC), New Orleans~(NO), San Antonio~(SA), Philadelphia~(Phi),  and Phoinex~(Pho). \vspace{-0em}} 
\label{tab:City-F1}
\end{table*}

\vspace{2mm} \noindent \textbf{Results.}
The results of the experiments are reported in Table~\ref{tab:city_pred}. Using content and style features, we were able to predict the location of a tweet more than 38\% of the time, an increase of almost 140\% in accuracy than the best random baseline, suggesting that both content and style features are predictive of the location a tweet is made. 
Likewise, using the POS and dialect features alone, the model achieves an accuracy of more than 35\%, substantially higher than the random baselines. Given that there are only four dialect features, this is indicative that the group information detected by the \citet{blodgett2016demographic} is informative. Similarly, the POS results are also high, indicating that there are unique combinations of POS patterns that appear in each location. Overall, the findings show that our subsample (GeoOLID) is representative of patterns found in the Unfiltered GeoOLID dataset.

\vspace{2mm} \noindent \textbf{Discussion.} 
\citet{blodgett2016demographic} show that the assumption that cities with large African American populations will have more text classified as AAE. Hence, as a simple robustness test, we use the tool provided by \citet{blodgett2016demographic} to correlate it with the demographic information of each city in our labeled GeoOLID dataset. Specifically, using the 2020 US Census data, we calculate the proportion of ``Black or African American alone'' (AA) and ``Hispanic or Latino'' (H/L) residents for each city. We also calculate the average African-American (AAS) and Hispanic (HS)  scores for each city using the tool from \citet{blodgett2016demographic}. Finally, we calculate the Pearson Correlation Coefficient (PCC) AA and AAS (and H/L and HS). The goal is to show that the findings found in \citet{blodgett2016demographic} hold on our GeoOLID dataset, i.e., their tool's scores correlate with minority populations. If they do, this provides further evidence that our dataset is representative of each location.


This correlation can be seen in Table \ref{tab:PCC}. Overall we find that there is significant correlation .565 ($p$ value: .028) between the two variables.
We find that cities like Baltimore, New Orleans and Detroit are more likely to have more AAE tweets (in the labeled GeoOLID dataset) then cities like Miami, Columbus, and New York. For the Hispanic group we also find a positive correlation but the finding is not significant. We also manually analyzed the dataset and found other features indicative of a relationship between demographics of the city and language use. For example, we found Spanish curse words appearing in text in cities with higher Hispanic populations in our dataset, e.g., ``Nationwide shutdown! pinché Cabron'' is an slightly modified tweet that was tagged in Phoenix, AZ.



\begin{table*}[t]
\centering
\resizebox{.85\textwidth}{!}{%
\begin{tabular}{lrrrrrrrrrrrrrrrr}
\toprule
       & \textbf{Bal} & \textbf{Chi} & \textbf{Col} & \textbf{Det} & \textbf{ElP} & \textbf{Hou} & \textbf{Ind} & \textbf{LA} & \textbf{Mem} & \textbf{Mia} &  \textbf{NO} & \textbf{NY} & \textbf{Phi} & \textbf{Pho} & \textbf{SA} & \textbf{AVG} \\ \midrule
\textbf{Linear SVM} & .187& .193&  .218& .233&  .239&  .211&  .167&  .172&  .247&  .152&  .194&  .151&  .191&  .247&  .220& .201 \\ \midrule
\textbf{BiLSTM}     & .111& .100&  .144& .149&  .154&  .142&  .115&  .119&  .142&  .085&  .118&  .104&  .116&  .154&  .143& .126 \\
\textbf{CNN}        & .124& .106&  .155& .168&  .168&  .159&  .112&  .137&  .167&  .091&  .115&  .104&  .126&  .174&  .164& .138  \\
\textbf{LSTM}       & .111& .092&  .133& .137&  .146&  .134&  .102&  .114&  .134&  .079&  .105&  .099&  .108&  .145&  .135& .118 \\
\textbf{BERT}       & \textbf{.105} & \textbf{.069} &  \textbf{.113} & \textbf{.087} &  \textbf{.075} &  \textbf{.075} &  \textbf{.068} &  \textbf{.078} &  \textbf{.118} &  \textbf{.059} &  \textbf{.086} &  \textbf{.070} &  \textbf{.091} &  \textbf{.115} &  \textbf{.097} & \textbf{.086} \\ \midrule
\textbf{AVG}        & .128 & .112&  .153& .155&  .156&  .145&  .113&  .124&  {.162} &  \cellcolor{red!25}.093 &  .124&  .106&  .126 &  \cellcolor{blue!25}.167 &  .152&  \\ \bottomrule
\end{tabular}%
}
\caption{False positive rate (FPR) for each model-city combination. The largest and smallest average values are shaded in \colorbox{blue!25}{blue} and \colorbox{red!25}{red}, respectively. The model with the \textit{smallest} (smaller is better) FPR is \textbf{bolded} for each city.\vspace{-0em}}
\label{tab:City-FPR}
\end{table*}

\begin{table}[t]
\centering
\resizebox{.5\linewidth}{!}{%
\begin{tabular}{lrr}
\toprule
       & \textbf{AA}         & \textbf{Hispanic}       \\ \midrule
\textbf{Spearman} & -.005 & -.103 \\
\textbf{PCC} & .102 & -.018\\ \midrule \midrule
\multicolumn{3}{c}{\textbf{AAE vs SAE Results}} \\ \midrule
\textbf{AAE FPR}  & \multicolumn{2}{c}{.154} (3392) \\ 
\textbf{SAE FPR}  & \multicolumn{2}{c}{.092} (5789) \\ \bottomrule
\end{tabular}}%
\caption{Correlation (Spearman and PCC) between the FPR scores and  AA and H/L population proportions of each city.\vspace{-1em}}
\label{tab:per-cor}
\end{table}

\subsection{Data Variation and Model Performance}
\label{sec:var}

Next, we measure how much offensive language detection performance can vary location-to-location.
Given the same model is applied to every city, ideally, we would have similar performance universally. However, if we see large variation in performance metrics and if the errors are caused by patterns also represented in the Unfiltered GeoOLID dataset, this is indicative of geographic performance disparities.

\vspace{2mm} \noindent \textbf{Methods.}
We train five different machine learning algorithms: Linear Support Vector Machine (Linear SVM), Long Short Term Memory (LSTM), Bidirectional LSTM (BiLSTM), Convolutional Neural Networks (CNN), and a Bidirectional Encoder Representations from Transformers (BERT). Each model is trained to classify Offensive and Non Offensive tweets using the OLID dataset.
One thing to note is for the BiLSTM, CNN and LSTM, we also measure the performance of the model across multiple word embeddings. Specifically, each deep learning model is trained using different variations of Glove, Google Word2Vec and Fasttext word embedding (See the Appendix, Table \ref{tab:words}, for a complete listing of the evaluated embeddings). 

For evaluation, We train multiple models on the OLID dataset using a 5-fold shuffle-split cross-validation procedure. Specifically, a model is trained on each training split of the OLID dataset, then it is applied to the GeoOLID dataset to calculate each city's model performance.  A 10\% portion of the OLID training split for each fold is used for hyperparameter selection.  This procedure has been used in prior work to ensure robust results in similar social media-related NLP studies~\cite{yin2017power,elejalde2017nature,samory2020characterizing}. 


\vspace{2mm} \noindent \textbf{Results.} In Table~\ref{tab:City-F1},we report the F1 of the OLID model applied to the GeoOLID dataset. Overall, we find substantial variation in model accuracy across the 15 cities. The average F1 for Houston (averaged across the non-random baselines) ranges from .588 (Houston) to .715 (Detroit), nearly 13\% percent absolute difference and 22\% relative difference. 
The CNN model achieves the best performance on average. Likewise, we find that CNN's best results are for the cities of Baltimore, Detroit, Memphis, and Philadelphia. Conversely, the CNN's worst results are found in Houston, Phoenix, Chicago, and New York.

Table~\ref{tab:City-FPR} reports the False Positive Rates (FPR) for each city. Again, we see large variation, ranging from .093 to .167, nearly an 80\% relative difference). On the other hand, we find that the best performing model is consistent across all cities. Hence, if a model performs better in Houston, it is likely it will perform better in Detroit. However, just because the model is better, the performance can be very low when compared to another location. Hence, decision-makers must carefully evaluate models based on the people impacted by them and not rely on evaluation metrics calculated on non-representative data before using the model. Finally, we report Accuracy results in the Appendix, Table \ref{tab:City-ACC}, which show improvement greater than chosen MDE thresholds.

Prior work by \citet{sap2019risk} show that offensive language detection models generate more false positives for text written in AAE. We evaluate this on our GeoOLID dataset following a similar strategy as \citet{sap2019risk}.
In Table~\ref{tab:per-cor}, we use the \citet{blodgett2016demographic} tool to identify AAE (African American English) and SAE (Standard American English) tweets in our GeoOLID dataset across all cities. When we calculate the false positive rate (FPR) across these two aggregate groups, we find similar conclusions to prior work~\cite{sap2019risk} suggesting that offensive language models generate more false positives on AAE text.

However, this does not mean that cities with larger African American populations will have larger FPRs. To test this using PCC and Spearman $\rho$, we correlate model performance (FPR) with the proportion of Black or African American and Hispanic/Latino residents (See Table~\ref{tab:PCC}) using US Census data. We find that there is weak to no correlation between FPR and minority population, which is surprising given AAE is correlated with population. There are two major reasons for this phenomena. First, AAE is not widely spoken. Even among African Americans, they do not always use an AAE dialect. Hence, if someone uses AAE sparingly, then the FPR on AAE will not accurately represent how the model will perform on text they write. Second, there are other factors that have a larger impact. In particular, this data is within the context of Covid-19. Hence, there are topics written with particular stances that the offensive langauge detection model is unable to handle, e.g., understanding the context of ``wuhan virus''.

Finally, we suggest that researchers should look at evaluating ``contextual language'', e.g., try to identify real people, ask them how they identify with regard to race and gender, then evaluate how models perform for them. This can provide insight into real bias issues and ground potential negative impact on real people. This idea fits with the narrative against treating groups as a monolith entities~\cite{tadjiogueu2014fifty, erving2021disrupting}.

\begin{figure} 
    \centering
    \includegraphics[width=.9\linewidth]{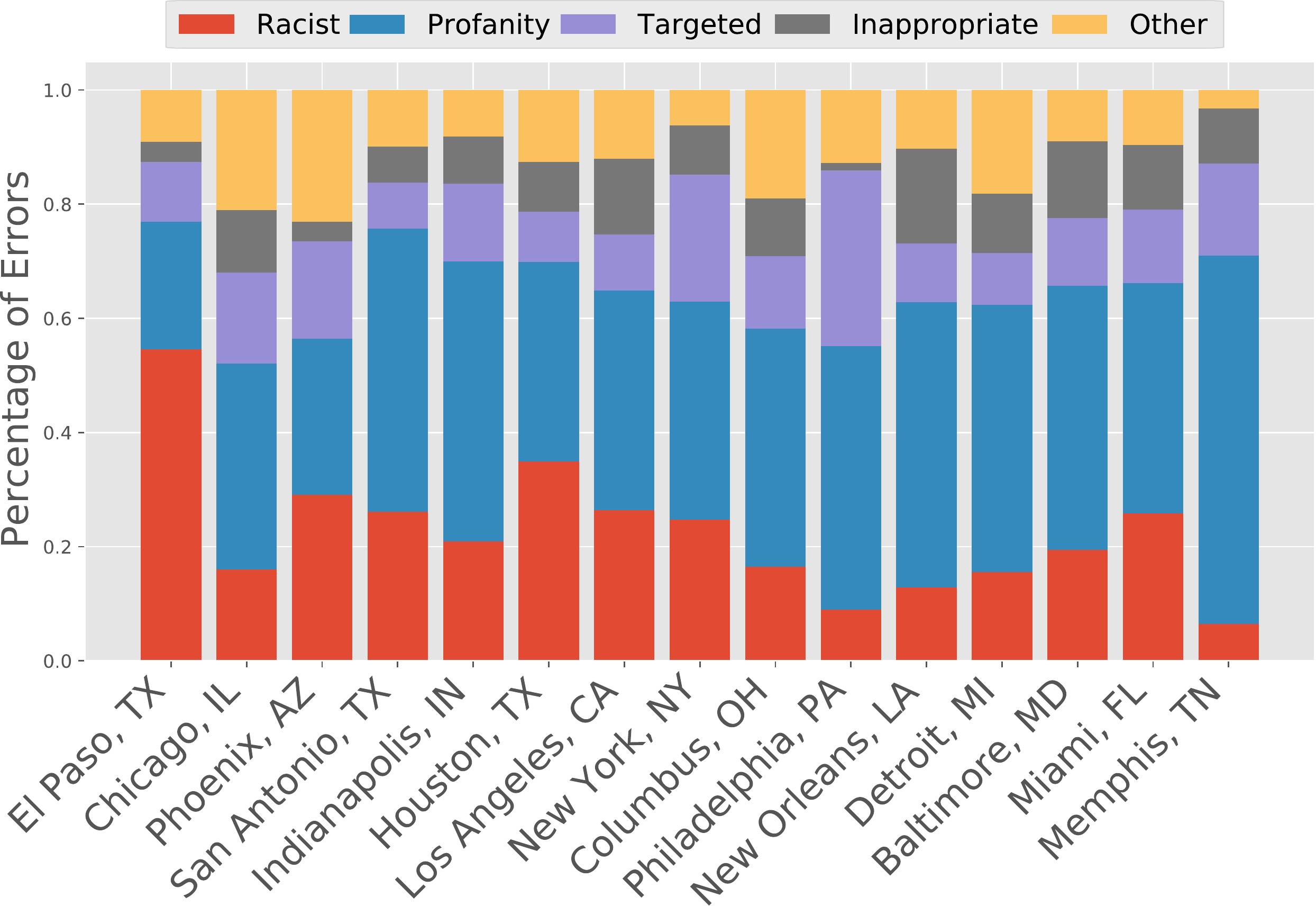}
    \caption{Manually coded false negatives per city.\vspace{-2em}}
    \label{fig:error_analysis_FN}
\end{figure}


\vspace{2mm} \noindent \textbf{Discussion.} We perform a comprehensive manual analysis on the false negatives made by the best model on the OLID dataset. Specifically, we performed a qualitative open coding procedure to categorize the false negatives into commonly occurring groups. We allowed categories and meanings to emerge from posts in somewhat of an open coding fashion~\cite{straussbasics}. We randomly sample up to 100 false negatives from each city, identifying the main categories. Next, a meeting was held where the main categories were discussed.

The final group of codes were identified as: Racist, Profanity, Targeted, Inappropriate, and Other. Racist was defined as a direct attack of mentioned of a race and/or ethnicity, Profanity as any sort of curse words, this could be in a hashtag or acronym. Targeted was defined as an attack on an individual, personal or group not associated to race or ethnicity, and finally Inappropriate is defined as any insensitive joke or sexual reference.

The results are summarized in Figure~\ref{fig:error_analysis_FN}. A few important observations can be made from this graph. For instance, we find a large proportion of false negatives in the racist category in border cities, or cities in close proximity to Mexico (e.g., El Paso, Phoenix, San Antonio, and Houston). The reason the false negatives occur is based on the stance, topic, and way of Racist writing we found to be common in the border regions. For instance, we found many issues where the model did not detect language that refers to migrants being part of a ``horde,'' meant to cause violence or destruction (this is common racist rhetoric at the time~\cite{finley2020immigrant}),  as being offensive.

\begin{figure}[t]
    \centering
    \includegraphics[width=.85\linewidth]{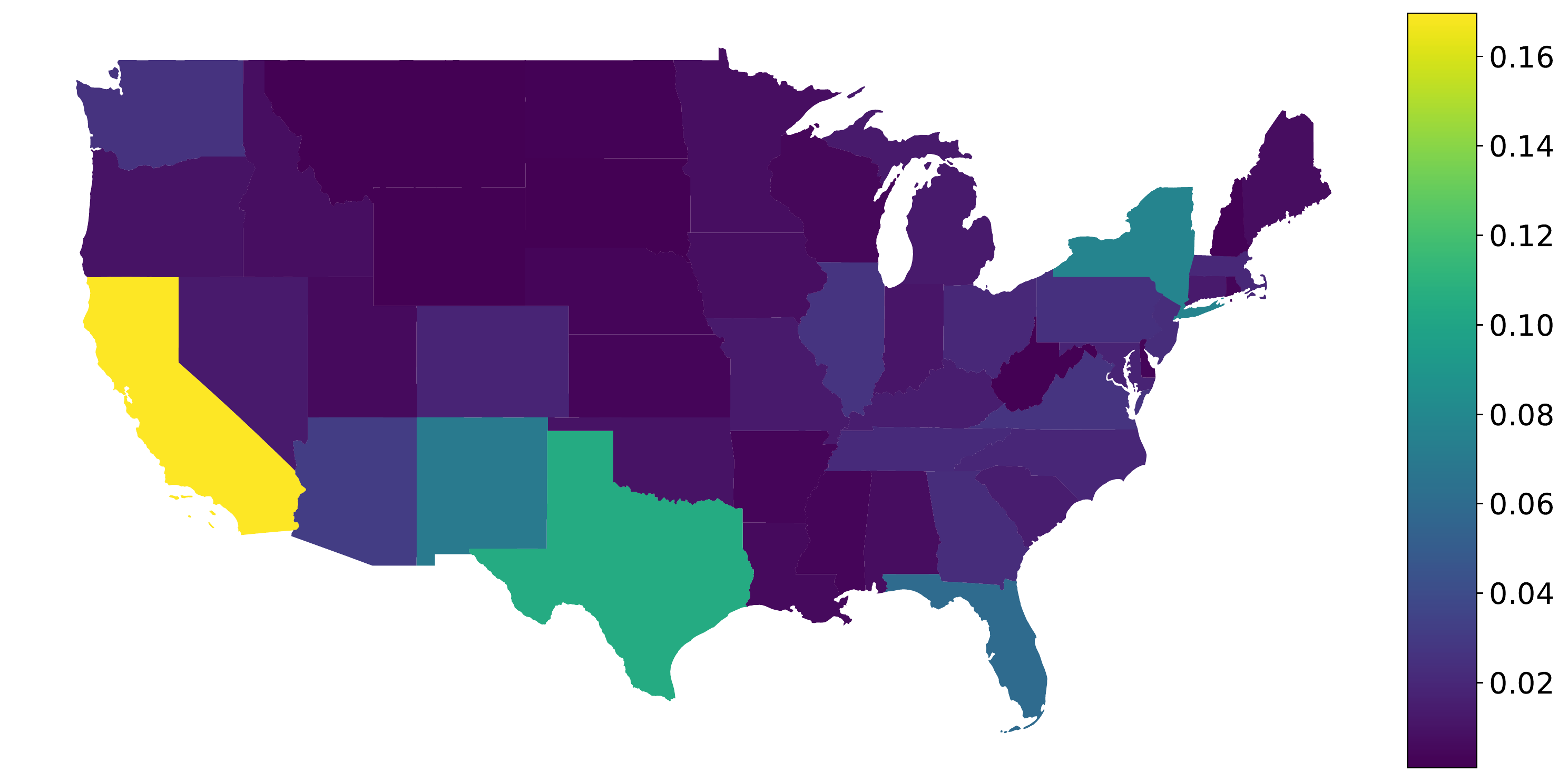}
    \caption{Proportion of border-related tweets in the ``Original Data'' for each state.\vspace{-2em}} 
    \label{fig:geo-map}

\end{figure}

We counted the number of border-related topics using a small set of search terms (e.g., ``border'', ``migrants'', ``immigrants'', and ``illegals'') in the Original Data. We plot the results in Figure~\ref{fig:geo-map}. We find that most of the border-related tweets are in states near Mexico (e.g., Texas, Arizona, New Mexico, and California). Hence, more false negatives caused by racist-categorized tweets about the border are more likely to be made in these cities, thus also increasing the likelihood of false negatives. Given the increase in border-related topics, this error is location-related.

Prior work has shown geographic differences in the use of swear words on social media~\cite{carey_2020,googlesites}. We also found morphological variants of curse words in different cities that caused false negatives. For example, in New Orleans,  Philadelphia, and Memphis there were many false negative tweets contain high percentages of Profanity due to multiple spellings of different swear words such as ``phucking'', ``effing'', ``mothafucka'', ``biatches''. 

\subsection{Geographic Similarities}
In this subsection, we analyze the correlation between the best-performing models in each city.

\paragraph{Methods.} We analyze the performance of the models trained and described in Section~\ref{sec:var}. Specifically, we compare the PCC between the Accuracy of each model applied to all cities. Intuitively, if the models for New York are sorted based on Accuracy, and the sorted order is the same as Phoenix, then the correlation would be one, showing a linear relationship. The more differences in sorted scores/models, the lower the performance (i.e., correlation). Overall, in this experiment, we rank every model along with the variants of models and compare every pair of cities rankings (i.e., each model trained with different word embeddings listed in the Appendix, Section~\ref{models-hypers} are treated as independent models). Intuitively, if correlations are very high, this could indicate that you could choose the best hyperparameters for city A and they would be the best for city B. From the main results in Table~\ref{tab:City-F1} we saw that the best model was the same across all cities. However, this is with substantial hyperparameter optimization. Are the parameters the same for each city?

\begin{figure}[]
    \centering
    \includegraphics[width=\linewidth]{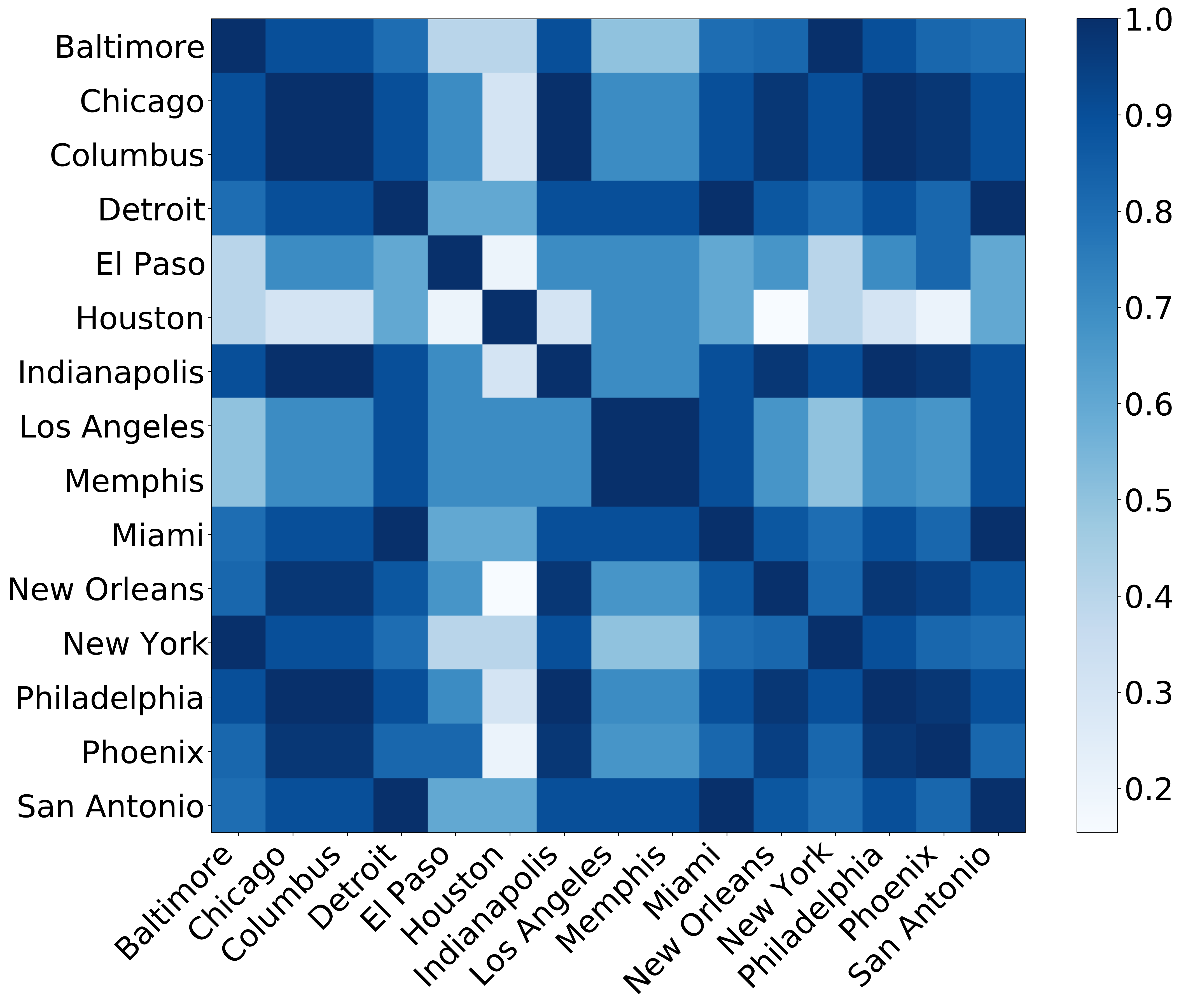}
    \caption{Model accuracy correlation between each pair of cities in GeoOLID.\vspace{-2em}}
    \label{fig:correlation-ACC}
\end{figure}

\paragraph{Results.} The results of the correlation analysis are shown in Figure~\ref{fig:correlation-ACC}. Overall, similar to variations in model performance across cities, we find that the similarity in model performance correlations can vary substantially city-to-city. For instance, the best models for Houston are substantially different from other cities, except for a few (e.g., Los Angeles). However, on further inspection, general architecture performance seems to be relatively similar across cities, e.g., the CNN model is the best on the OLID dataset and for most cities. Much of the variation comes from hyperparameter choice or pretrained embedding choice (with more than 10\% in Accuracy between the best and worst embeddings). The best embeddings can be substantially different city-to-city. This result suggests that choosing the best hyperparameters based on a small subset of data (e.g., from one city) is not optimal for each location, which can result in further performance disparities. 

\paragraph{Discussion.}
The results do provide us with a potential research avenue. An interesting question that could be explores is if we train a model with many hyperparameter options on a dataset, is it possible to predict which model to deploy in a given region? There has been some work in predicting model perform~\cite{elsahar2019annotate}. Hence, it would be interesting to expand that to predict the best hyperparameters.

\section{Conclusion}
We provide a comprehensive analysis of performance disparities of offensive language detection models. Also, we introduce a novel dataset that provides more than 14 thousand examples for analysis of geographical differences in model performance. The study points to the importance of geographically sensitive NLP, where the impact and performance of NLP models are analyzed for specific geographical regions, or even micro communities within a city. Moreover, finding regions where models perform poorly on can also provide unique testbeds as ``hard test cases'' similar to recent work on adversarial examples~\cite{zhang2019generating}.

\section*{Acknowledgements}

This material is based upon work supported by the National Science Foundation (NSF) under Grant~No. 1947697 and NSF award No. 2145357.

\bibliographystyle{acl_natbib}

\bibliography{custom}

\appendix

\section{Appendix}

\subsection{Word Embeddings}

In Table~\ref{tab:words}, we link to the publicly available word embeddings we use in our experiments. We test three models: SkipGram, GLOVE, and FastText. We also explore different embeddings sizes, ranging for 25 dimensions to 30. Moreover, we explore embeddings trained on different corpora, ranging from biomedical text (PubMed) to social media data (Twitter). The best embeddings are chosen based on the OLID validation dataset for all reported results in the main manuscript.

\begin{table*}[b]
\centering
\resizebox{\textwidth}{!}{%
\begin{tabular}{@{}lrrp{8cm}}
\toprule
Model & Data Source & Dimension & Link \\ \midrule
SkipGram & Google News & 300 & \url{https://docs.google.com/file/d/0B7XkCwpI5KDYaDBDQm1tZGNDRHc/edit?usp=sharing} \\
SkipGram & PubMed & 200 & \url{http://evexdb.org/pmresources/vec-space-models/PubMed-w2v.bin} \\
SkipGram & PubMed Central & 200 & \url{http://evexdb.org/pmresources/vec-space-models/PMC-w2v.bin} \\
SkipGram & PubMed and PubMed Central & 200 & \url{http://evexdb.org/pmresources/vec-space-models/PubMed-and-PMC-w2v.bin} \\
SkipGram & Wikipedia, PubMed, and PubMed Central & 200 & \url{http://evexdb.org/pmresources/vec-space-models/wikipedia-pubmed-and-PMC-w2v.bin} \\
GLOVE & Twitter & 25 & \url{http://nlp.stanford.edu/data/glove.twitter.27B.zip} \\
GLOVE & Twitter & 50 & \url{http://nlp.stanford.edu/data/glove.twitter.27B.zip} \\
GLOVE & Twitter & 100 & \url{http://nlp.stanford.edu/data/glove.twitter.27B.zip} \\
GLOVE & Twitter & 200 & \url{http://nlp.stanford.edu/data/glove.twitter.27B.zip} \\
GLOVE & Wikipedia 2014 and Gigaword 5  & 50 & \url{http://nlp.stanford.edu/data/glove.6B.zip} \\
GLOVE & Wikipedia 2014 and Gigaword 5 & 100 & \url{http://nlp.stanford.edu/data/glove.6B.zip} \\
GLOVE & Wikipedia 2014 and Gigaword 5 & 200 & \url{http://nlp.stanford.edu/data/glove.6B.zip} \\
GLOVE & Wikipedia 2014 and Gigaword 5 & 300 & \url{http://nlp.stanford.edu/data/glove.6B.zip} \\
GLOVE & Common Crawl V1 & 300 & \url{http://nlp.stanford.edu/data/glove.42B.300d.zip} \\
GLOVE & Common Crawl V2 & 300 & \url{http://nlp.stanford.edu/data/glove.840B.300d.zip}      \\
FastText & Wikipedia 2017, UMBC webbase corpus, and statmt.org news dataset & 300 & \url{https://dl.fbaipublicfiles.com/fasttext/vectors-english/wiki-news-300d-1M.vec.zip} \\
FastText & Common Crawl & 300 & \url{https://dl.fbaipublicfiles.com/fasttext/vectors-english/crawl-300d-2M.vec.zip} \\ \bottomrule
\end{tabular}}
\caption{List of word embeddings we use in our experiments.}
\label{tab:words}
\end{table*}

\subsection{Model Hyper-parameters}\label{models-hypers}

In this Section, we report the best hyperparmeters for each model. For the linear models we also report the best TF-IDF settings from the scikit-learn package.

\vspace{1mm} \vspace{2mm} \noindent \textbf{TF-IDF:}
\begin{itemize}[topsep=0pt]
    \itemsep-.5em
      \item sublinear tf: True
      \item min df: 5
      \item norm: l2
      \item encoding: latin-1
      \item ngram range: (1,2)
      \item stop words: english
\end{itemize}

\vspace{1mm} \vspace{2mm} \noindent \textbf{Linear SVM:}
\begin{itemize}[topsep=0pt]
    \itemsep-.5em
    \item penalty: l2
     \item C: 1.0
\end{itemize}

\vspace{1mm} \vspace{2mm} \noindent \textbf{CNN:}
\begin{itemize}[topsep=0pt]
    \itemsep-.5em
      \item max words: 10000
      \item max sequence length: 125
      \item drop: .2
      \item batch size: 512
      \item epochs: 30
      \item filter sizes: 3,4,5
      \item num filters: 512
      \item early stopping: 5 iterations 
\end{itemize}

\vspace{1mm} \vspace{2mm} \noindent \textbf{LSTM:}
\begin{itemize}[topsep=0pt]
    \itemsep-.5em
      \item max words: 10000
      \item max sequence length: 125
      \item drop: .2
      \item batch size: 128
      \item epochs: 30
      \item num filters: 512
      \item hidden layers: 1
      \item early stopping: 5 iterations 
\end{itemize}

\vspace{1mm} \vspace{2mm} \noindent \textbf{BiLSTM:}
\begin{itemize}[topsep=0pt]
    \itemsep-.5em
      \item max words: 10000
      \item max sequence length: 125
      \item drop: .2
      \item batch size: 128
      \item epochs: 30
      \item num filters: 512
      \item hidden layers: 1
      \item early stopping: 5 iterations 
\end{itemize}

\vspace{1mm} \vspace{2mm} \noindent \textbf{BERT:}
\begin{itemize}[topsep=0pt]
 \itemsep-.5em
      \item tokenizer : bert-base-cased
      \item model : bert-base-cased
      \item dropout : .2
      \item max length : 128
      \item epochs : 50
      \item batch size : 64
      \item fine tuned : after 5 epochs
      \item early stopping : 5 iterations 
\end{itemize}

\begin{table}[t]
\centering
\resizebox{.9\linewidth}{!}{%
\begin{tabular}{lll}
\toprule
\textbf{Word Embedding} & \textbf{F1} & \textbf{Accuracy} \\ \toprule
\textbf{BiLSTM}  & & \\ \midrule
FASTTEXT\_en\_300 & .580 & .760 \\
GLOVE\_twitter\_27B\_100d & .627 & .785 \\
GLOVE\_twitter\_27B\_50d & .5834 & .764 \\
GLOVE\_wiki\_42B\_300d & .645 & .793 \\
GLOVE\_wiki\_6B\_100d & .600 & .771 \\
GLOVE\_wiki\_6B\_200d & .605 & .778 \\
GLOVE\_wiki\_6B\_300d & .631 & .783\\
GLOVE\_wiki\_6B\_50d & .586 & .768 \\
GLOVE\_wiki\_840B\_300d & .631 & .787 \\
W2V\_GoogleNews & .616 & .781 \\
W2V\_PMC & .488 & .730 \\
W2V\_PubMed\_PMC & .514 & .738 \\
W2V\_PubMed & .402 & .704 \\ \toprule
\textbf{LSTM} & & \\ \toprule
FASTTEXT\_en\_300 & .524 & .749 \\
GLOVE\_twitter\_27B\_100d & .618 & .782 \\
GLOVE\_twitter\_27B\_50d & .591 & .770 \\
GLOVE\_wiki\_42B\_300d & .619 & .790 \\
GLOVE\_wiki\_6B\_100d & .607 & .774 \\
GLOVE\_wiki\_6B\_200d & .616 & .781 \\
GLOVE\_wiki\_6B\_300d & .609 & .782 \\
GLOVE\_wiki\_6B\_50d & .577 & .762 \\
GLOVE\_wiki\_840B\_300d & .624 & .788 \\
W2V\_GoogleNews & .602 & .779 \\
W2V\_PMC & .456 & .720 \\
W2V\_PubMed\_PMC & .495 & .730 \\
W2V\_PubMed & .348 & .701 \\ \toprule
\textbf{CNN} & & \\ \toprule
FASTTEXT\_en\_300 & .611 & .778 \\
GLOVE\_twitter\_27B\_100d & .657 & .792 \\
GLOVE\_twitter\_27B\_50d & .635 & .788 \\
GLOVE\_wiki\_42B\_300d & .642 & .793 \\
GLOVE\_wiki\_6B\_100d & .621 & .779 \\
GLOVE\_wiki\_6B\_200d & .621 & .786 \\
GLOVE\_wiki\_6B\_300d & .621 & .785 \\
GLOVE\_wiki\_6B\_50d & .612 & .775 \\
GLOVE\_wiki\_840B\_300d & .648 & .794 \\
W2V\_GoogleNews & .638 & .789 \\
W2V\_PMC & .520 & .738 \\
W2V\_PubMed\_PMC & .541 & .743 \\
W2V\_PubMed & .461 & .718 \\ \bottomrule
\end{tabular}
}
\caption{Word Embedding Performance for Deep Learning Models}
\label{tab:wordembedding-deep}
\end{table}

\begin{table}[t]
\centering
\resizebox{.7\linewidth}{!}{%
\begin{tabular}{lrrrr}
\toprule
           & \textbf{Prec.} & \textbf{Rec.} & \textbf{F1} & \textbf{Acc.} \\ \midrule
           \multicolumn{5}{c}{\textbf{Random Baselines}} \\ \midrule
\textbf{Stratified} &   .324 & .348 & .336 & .553          \\
\textbf{Uniform}    &     .321 & .505 & .392 & .493         \\ \midrule \midrule
\multicolumn{5}{c}{\textbf{Machine Learning Models}} \\ \midrule
\textbf{Linear SVM} &        .643 & .505       &  .566  & .744          \\ \midrule
\textbf{BiLSTM}     &       .754    &   .551     &  .631  &    .783     \\
\textbf{CNN}        &       .721 & .603       & .657   &    .792      \\
\textbf{LSTM}       &     .768      &   .527     & .624  &      .788     \\
\textbf{BERT}        & .652 & .555 & .592   &  .752        \\ \bottomrule
\end{tabular}%
}
\caption{OLID Results\vspace{-1em}}
\label{tab:OLID-results}
\end{table}

\subsection{OLID Results}

We report the OLID results for each model (Linear SVM, CNN, LSTM, BiLSTM, and BERT) in Table~\ref{tab:OLID-results}. Interestingly, we find that the CNN model outperforms other methods, including the LSTM-based models and BERT. For instance, the CNN's F1 is more than 2\% higher than the LSTM and BiLSTM models. Moreover, it is more than 6\% higher than BERT. We also find that all methods outperform the traditional machine learning models (Linear SVM), with the CNN outperforming the Linear SVM by nearly 9\% F1 and nearly 5\% in Accuracy. The results support the results of the main paper with the CNN model generalizing better than other techniques.

Next, in Table~\ref{tab:wordembedding-deep} we report the performance of the CNN, LSTM, and BiLSTM models trained using different embeddings. Overall, we see variation across which embeddings result in teh best F1 score for each model, with wiki\_42B\_300d resulting in the highest F1 for the BiLSTM, wiki\_840B\_300d resulting in the best results for the LSTM, and GLOVE\_twitter\_27B\_100d. This finding is similar to the results for H3 in the main paper, where embedding choice can vary city-to-city. We also find that it can vary model-to-model, which is also supported in \citet{rios2020empirical}.

\subsection{Accuracy Power Analysis}
\begin{figure}
    \centering
    \includegraphics[width=\linewidth]{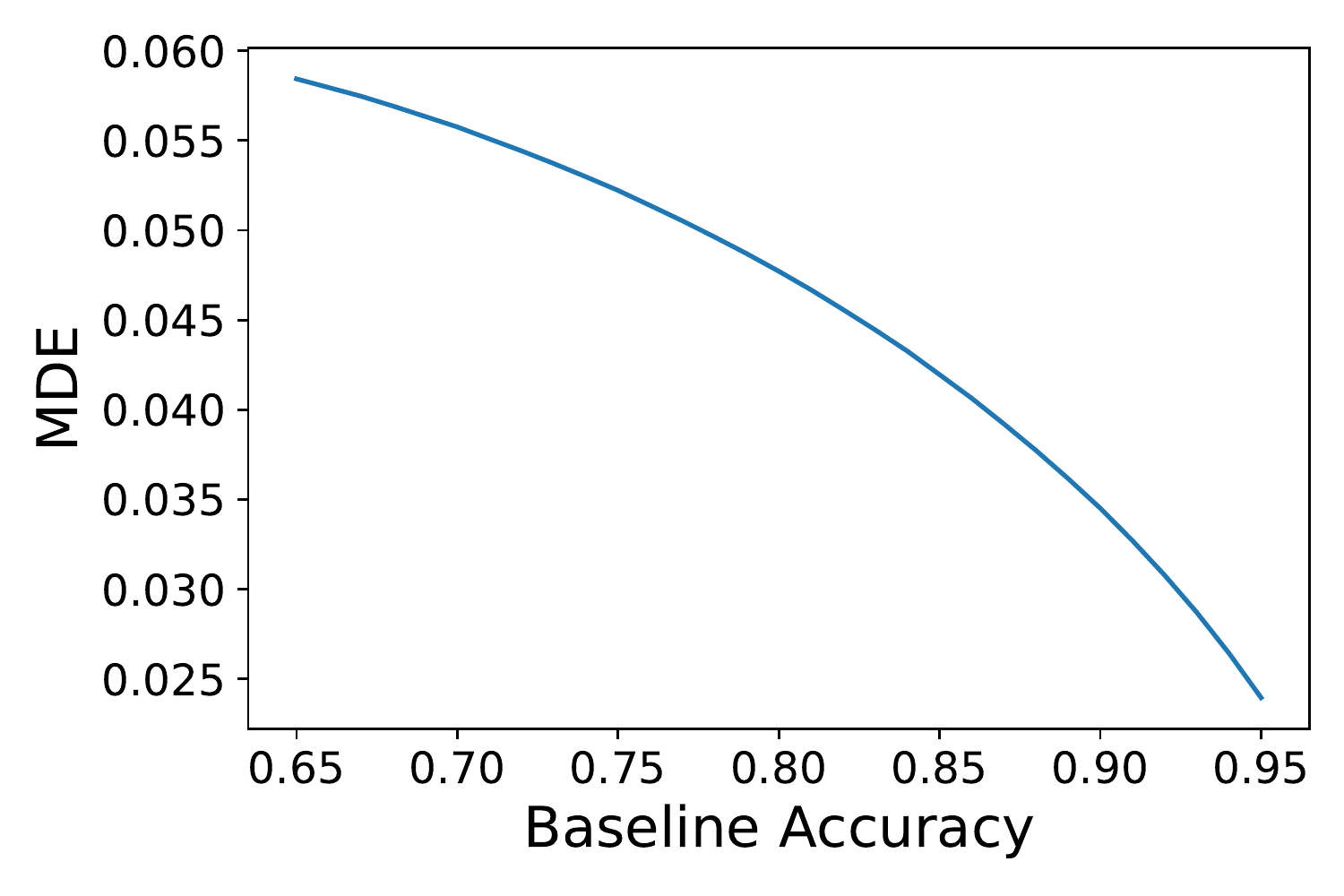}
    \caption{MDE given different baseline accuracy assumptions and a power of 80\%.}
    \label{fig:acc-power}
\end{figure}

In Figure~\ref{fig:acc-power}, we report the MDE~\cite{card2020little} for Accuracy assuming different baseline scores and a power of 80\%. For instance, if the baseline achieves an accuracy of .95, then we would need to see any improvement/difference of around .025 for it to be significant. Likewise, if the accuracy is around .65, then we need an improvement of nearly .06 for it to be significant. Intuitively, the more accurate the results, the smaller the improvement can be for it be significant.

\subsection{Accuracy Scores per City}

In Table~\ref{tab:City-ACC}, we report the OLID model accuracy for each city. Overall, we find substantial variation in model accuracy across the 15 cities. The Linear SVM classifier ranges from .704 to .822, resulting in around a 12\% difference in accuracy between Phoenix and Miami. Similar findings can be seen with the other models like CNN and BERT having a up to a 10\% difference. Furthermore, given the MDE of around 5\% for each city depending on the baseline score, we find that many of the differences are significant.

\begin{table*}[t]
\centering
\resizebox{\textwidth}{!}{%
\begin{tabular}{lrrrrrrrrrrrrrrrr}
\toprule
       & \textbf{Bal} & \textbf{Chi} & \textbf{Col} & \textbf{Det} & \textbf{ElP} & \textbf{Hou} & \textbf{Ind} & \textbf{LA} & \textbf{Mem} & \textbf{Mia} &  \textbf{NO} & \textbf{NY} & \textbf{Phi} & \textbf{Pho} & \textbf{SA} & \textbf{AVG} \\ \midrule
\textbf{Stratified} & .555 & .555 & .550 & .536 & .536 & .570 & .577 & .567 & .521 & .592 & .553 & .588 & .567 & .544 & .564 &  .558 \\ \midrule \midrule
\textbf{Linear SVM} & .779          & .745        & .751         & .761        & .694        & .724        & .748             & .776            & .752        & .822      & .787        & .794         & .771             & .704        & .740        &   .757 \\ \midrule
\textbf{BiLSTM}    & .834 & .809 & \textbf{.799} & .803 & .757 & .774 & .809 & .824 & \textbf{.818} & .861 & .835 & .842 & .833 & .768        & .783 &   .809  \\
\textbf{CNN} & \textbf{.843} & \textbf{.820} & .792 & .823 & .747 & .773 & \textbf{.819} & .805 & .814 & .851 & \textbf{.842} & \textbf{.849} & .849 & .760 & \textbf{.788} &   .811  \\
\textbf{LSTM}       & .832 & .814 & .790 & \textbf{.834} & \textbf{.758} & .790 & .817 & \textbf{.829} & .810 & \textbf{.873}      & .837        & .834         & \textbf{.850}             & \textbf{.772}        & .783       &    \textbf{.815} \\
\textbf{BERT}       & .786             &  .800          & .788            & .785           & .755           & \textbf{.791}           & .790                & .809              & .761           & .848         & .785           & .816            & .803                & .747           & .771         &     .789 \\ \midrule
\textbf{AVG} & .815 & \cellcolor{yellow!75}.798 & \cellcolor{yellow!75}.784 & \cellcolor{yellow!75}.801 & \cellcolor{yellow!75} \textbf{.742} & \cellcolor{yellow!75}.770 & \cellcolor{yellow!75}.797 & \cellcolor{yellow!75}.809 & \cellcolor{yellow!75}.791 & \textbf{.851} & .817 & .827 & .821 & \cellcolor{yellow!75}.750 & \cellcolor{yellow!75}.773 & \cellcolor{yellow!75}.796 \\ \bottomrule
\end{tabular}%
}
\caption{Accuracy for each city in the GeoOLID dataset. In the bottom row (i.e., the average across all machine learning models), we mark the cities that have an average accuracy difference greater than or equal to the MDE compared to the city with the highest average accuracy.\vspace{-1em}}
\label{tab:City-ACC}
\end{table*}

\end{document}